\documentclass{llncs}
\usepackage{lineno,hyperref}
\usepackage{algorithm,algpseudocode}
\usepackage{amsmath}
\usepackage{algorithmicx}
\usepackage{verbatim}
\usepackage{blindtext}
\usepackage{algpseudocode}
\usepackage{acro}
\usepackage{doi}
\usepackage[normalem]{ulem}
\usepackage{graphicx}
\usepackage{epstopdf}
\usepackage{emf}
\usepackage{subcaption}
\usepackage{booktabs}
\usepackage{tabularx}
\usepackage{ltablex,lscape}
\usepackage{afterpage}
\usepackage{array}
\usepackage{float}
\usepackage{multirow}
\usepackage{multicol}
\usepackage{rotating}
\usepackage{enumitem}
\usepackage{pifont}
\usepackage{bigstrut}
\usepackage{color,soul}
\usepackage{booktabs}
\usepackage{url}
\usepackage{cleveref}
\usepackage{nicefrac} % For 1/2
\usepackage[utf8]{inputenc}
\usepackage{dingbat}
\modulolinenumbers[5]
\DeclareUnicodeCharacter{0301}{\'{e}}
\newcolumntype{H}{>{\iffalse}c<{\fi}@{}}
\DeclareUnicodeCharacter{2212}{-}

\let\oldtabular\tabular
\renewcommand{\tabular}{\footnotesize\oldtabular}

\let\oldalg\algorithmic
\renewcommand{\algorithmic}{\footnotesize\raggedright\oldalg}

\begin{document}

\captionsetup[table]{
	labelsep = newline,
	labelfont = bf,
	textfont = footnotesize,
	name = Table,
	justification=justified,
	singlelinecheck=false,%%%%%%% a single line is centered by default
	skip = \medskipamount}

\newcolumntype{L}[1]{>{\raggedright\arraybackslash}p{#1}}
\newcolumntype{C}[1]{>{\centering\arraybackslash}p{#1}}
\newcolumntype{R}[1]{>{\raggedleft\arraybackslash}p{#1}}

\begin{frontmatter}

\title{An Efficient Binary Harris Hawks Optimization based on Quantum SVM for Cancer Classification Tasks}

%% or include affiliations in footnotes:a
\author{Essam H. Houssein\inst{1} \and Zainab Abohashima\inst{2} \and Mohamed Elhoseny\inst{3} \and Waleed M. Mohamed\inst{1}}

\institute{Faculty of Computers and Information, Minia University, Egypt. \and Faculty of Computer Science, Nahda University, Beni-Suef, Egypt.\and Department of Computer Science, American University in the Emirates, Dubai, UAE.}

%E-mail: essam.halim@mu.edu.eg
%E-mail: waleedmakram@minia.edu.eg

\titlerunning{An Efficient Binary Harris Hawks Optimization based on Quantum SVM for Cancer Classification Tasks}
\maketitle

\begin{abstract}
\footnotesize{
Cancer classification based on gene expression increases early diagnosis and recovery, but high-dimensional genes with a small number of samples are considered the major challenge in cancer classification tasks.  Gene selection is a critical step in large-dimensional features to select the most informative and relevant subset of genes. The Gene selection problem utilizes metaheuristic techniques to extract the optimal subset of genes in microarray cancer datasets, such as particle swarm optimization (PSO), Harris hawks optimization (HHO), and grey wolf optimizer (GWO). Binary Harris hawk optimization (BHHO) technique, which mimics the behavior of the cooperative action of Harris hawks in nature, is recently proposed as one of these techniques.  The biggest challenge of quantum hardware is the limited number of qubits, which restricts the use of quantum devices in real applications. The principal component analysis (PCA) is applied to reduce the selected genes to match the qubit numbers.  In this work, a new hybrid quantum-kernel support vector machine (QKSVM) with BHHO called BHHO–PCA–QKSVM for cancer classification on a quantum simulator. This study aims to improve the microarray cancer prediction performance with the quantum kernel estimation based on the informative genes by BHHO. The quantum computer is used for estimation the kernel with the training data of the reduced genes and generation of the quantum kernel matrix.  Besides, the classical computer is used for drawing the support vectors based on the quantum kernel matrix and applying the prediction stage. The colon and breast microarray datasets are used for evaluating the proposed approach performance with all genes and the selected genes. The proposed model enhances the overall performance of the two datasets.  Also, the final results of the proposed model compared with different quantum feature maps (kernels) and classical radial basis function (RBF) kernel.}
\end{abstract}
\end{frontmatter}

\section{Introduction}
\label{Sec:Introduction}
Quantum computing recently emerged with machine learning (ML) to introduce a new field, called, quantum machine learning (QML) \cite{biamonte2017quantum}. Quantum computing is applied to enhance classical ML. QML is introduced in many concepts, including QML, quantum-hybrid ML, and quantum-inspired ML \cite{abohashima2020classification}.Various ML algorithms have been proposed in quantum versions, such as quantum k-means clustering \cite{lloyd2013quantum}, quantum nearest-neighbor \cite{wiebe2018quantum}, and decision tree classifier \cite{lu2014quantum}. The QSVM is introduced in many versions \cite{li2015experimental,rebentrost2014quantum,willsch2020support}. In this work, we focus on a quantum kernel-enhanced SVM. Kernel functions for ML are commonly used in pattern recognition and classification tasks. Kernel estimation has become expensive and difficult for classical devices with the data size increase. Quantum devices can easily estimate the kernel function with large-dimensional vectors; however, the current quantum hardware is limited by the available number of qubits.

With the high-dimensionality genes in microarray cancer data, the microarray datasets contain excessive and insignificant genes. So, the overall accuracy of the ML model can get largely poor. Therefore, feature selection is a vital step for selecting valuable genes and enhancing performance. The filter and wrapper methods are the main feature selection techniques for gene cancer data. The wrapper method is more significant in feature selection \cite{hussain2021efficient}. The gene selection process can be split as diversification of all available solutions. Each solution is evaluated by the ML algorithm (e.g., SVM) \cite{al2017examining,houssein2021hybrid2}.

Several meta-heuristic algorithms based on feature selection have been proposed \cite{hussien2019s,neggaz2020efficient}. Harris hawk optimization (HHO) is recently presented to solve optimization problems. HHO has a proven outstanding performance in several real-time optimization compared to other metaheuristic algorithms \cite{heidari2019harris} and is applied in many applications \cite{houssein2020hybrid,hashim2020modified,houssein2020novel}. In addition, in large-scale wireless sensor networks, the optimal sink node placement based on HHO is proposed \cite{houssein2020optimal}.
Meanwhile, ML in the healthcare domain has recently made progress. ML plays a critical role in solving most problems related to the healthcare field, including diagnosis and classification of medical data, extraction patterns in data, and detection of chronic diseases to avoid the risk of diseases. The most common chronic diseases are cancer, stroke, diabetes, and kidney and heart diseases. The selection of informative genes and the classification of high-dimensional features are some of the most important research topics. The number of genes in microarray data has a large dimensionality. Thus, the feature selection step is necessary with microarray datasets. This approach aims to
\begin{itemize}
\item Enhance the QKSVM performance with the quantum kernel estimation based on the all/selected genes by BHHO .
\item Investigate the proposed approach performance with various quantum kernels, and
\item Open the way for researchers to apply QML algorithms with real-world problems.
\end{itemize}
The key contributions of this work are as follows:
\begin{itemize}
\item A hybrid QKSVM approach with PCA and BHHO-based feature selection for microarray cancer detection, called BHHO–PCA–QKSVM, is proposed.
\item High-dimensional medical data are classified on a quantum simulator computer.
\item The performance of microarray cancer classification with different quantum kernels is demonstrated.
\end{itemize}

The remainder of this paper is structured as follows: \ref{Sec:Proposed} introduces the hybrid approach BHHO–PCA–QKSVM in detail; \ref{Sec:Results} discusses the experimental results and provides a comparative performance with different quantum kernels; and finally, section \ref{Sec:Con} concludes this work.

\section{The Proposed Approach}
\label{Sec:Proposed}
\begin{figure}[h]
	\centering
	\includegraphics
	[scale=0.7]{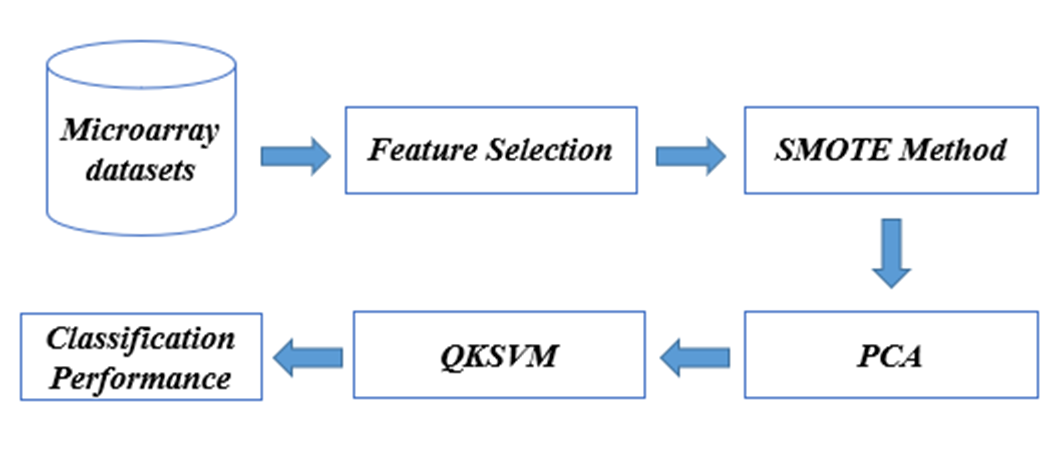} 
	\caption{Block diagram of the proposed approach.}
	\label{Fig:F1}
\end{figure}

Due to the limited quantum hardware resources, the advantage of quantum SVM is not clear enough in real applications \cite{phillipson2020three}. The proposed hybrid approach contains four steps: first, the BHHO selects informative genes; second, the PCA reduces the number of selected genes compatible with the number of qubits in the quantum device; third, the SMOTE method is used to handle the imbalanced classification problem; and fourth, the quantum kernel SVM is applied for the gene classification. Figure \ref{Fig:F1} shows the block diagram of the proposed approach.

\subsection{Harris Hawks Optimization}
The HHO is a new technique inspired by the chasing style proposed in 2019. The substantial idea of the HHO is based on mimicking the surprise pounce behavior. The surprise pounce is the cooperative action of Harris’ hawks in nature. The HHO consists of two phases: exploration and exploitation phases. In the exploration stage, the Harris hawks monitor and detect the victim (i.e., rabbit) using their sharp eyes. Harris hawks randomly settle in various spaces to attach to their prey. Mathematically, the exploration stage is defined as:

\begin{equation}
Z(t+1)=\left\{\begin{array}{ll}
Z_{k}(t)-r_{1}\left|Z_{k}(t)-2 r_{2} Z(t)\right|&e\geq 0.5 \\
\left(Z_{r}(t)-Z_{m}(t)\right)-r_{3}\left(lb+r_{4}(ub-lb)\right)& e<0.5
\end{array}\right.
\end{equation}
Where $Z$ is the Harris hawk location, and $t$ is the current iteration. The first law of the equation is to select the hawk location randomly. In the second law of the equation, $Z_r$ is the best solution (victim location); $Z_m$ is the mean hawk group location calculated from Eq.\ref{Eq: 2}, $r_1, r_2, r_3, r_4$, and $e$ are the numbers at random [0,1]. ub and lb are the maximum and minimum boundaries, respectively.
\begin{equation}
\label{Eq: 2}
Z_{m}(t)=\frac{1}{N} \sum_{n=1}^{N} Z_{n}(t)
\end{equation}
Where $N$ is the Harris hawks number, and $X_n$  is the hawk number in the group. The change of hawk action relies on the escaping energy of the victim ($E$) to transform into the exploitation stage. This can be modeled as:
\begin{equation}
E=2 E_{0}\left(1-\frac{t}{T}\right), E_{0}=2 r-1
\end{equation}
Where $E_0$ is the random initial energy in [-1, 1]; $T$ is the total number of iterations;  $r$ is number at random [0, 1], and $t$ is the current iteration.
In the exploitation stage, the hawk location is varied based on the victim’s energy (escaping energy). This stage is performed following the four main steps of the smooth blockade, hard blockade, smooth blockade with a progressive quick pounce, and hard blockade with a progressive quick pounce.
\subsubsection{Binary Harris Hawks Optimization}
Jingwei et al. \cite{too2019new} presented a new version of HHO for feature selection tasks, called binary HHO (BHHO). The BHHO algorithm is applied  to the transfer function (e.g., S-shaped) to convert the HHO to binary feature selection technique. \cite{saremi2015important}. The Harris hawk location is updated in two points. The binary HHO updates the current location $Z_{i}^{d}(t)$ of the Harris hawk into a new location $\Delta Z_{i}^{d}(t+1)$. The using V-/S-shaped function is used to apply the probability form of the HHO algorithm. The new hawk location using S-shaped functions is defined as follows:
\begin{equation}
Z_{i}^{d}(t+1)=\left\{\begin{array}{ll}
1 & \text { if rand }(0,1)<T\left(\Delta Z_{i}^{d}(t+1)\right) \\
0 & \text { otherwise }
\end{array}\right.
\end{equation}
Where $\left\{\begin{array}{l}\text { Z is the hawk location } \\ \text { i is the hawk number in the group } \\ \text { d is the dimension } \\ \text { t is the current iteration }\end{array}\right.$ \\
The new location of the hawk by using V-shaped functions can be modeled as follows:
\begin{equation}
Z_{i}^{d}(t+1)=\left\{\begin{array}{c}
\neg X_{i}^{d}(t) \text { if rand }(0,1)<T\left(\Delta Z_{i}^{d}(t+1)\right) \\
Z_{i}^{d}(t) \text { otherwise }
\end{array}\right.
\end{equation}
Where $\left\{\begin{array}{l}\mathrm{Z} \text { is the hawk location } \\ \mathrm{i} \text { is the hawk number in the group } \\ \mathrm{t} \text { is the current iteration } \\ \mathrm{d} \text { is the dimension } \\ \neg \mathrm{X} \text { is the complement of } \mathrm{X}\end{array}\right.$
\subsection{Synthetic Minority Oversampling TEchnique}
The synthetic minority oversampling technique (SMOTE) is a method to handle the imbalanced class problem in large dimensions’ data \cite{fernandez2018smote}. The key idea of SMOTE is to increase the number of observations in the minority class to be equal to the sample numbers in the majority class. In the colon dataset, the minority class (normal) has 22 samples. After using the SMOTE methods, the normal class has 31 samples. In the breast dataset, the minority class has 10 normal samples. After applying the SMOTE method, the normal class 64 samples (as seen in Table \ref{Tbl:T1}).
\subsection{Principal Component Analysis}
Currently, one of the challenges of QML is the limited number of qubits, which restricts the use of quantum devices in applications with QML algorithms \cite{abohashima2020classification,chen2021end}. Therefore, using a quantum computer with complex large dimensionality data is still difficult. In this study, the PCA method is used to transform the selected gene number to match the available number of qubits, which still retains most of the gene information of the genes.
\subsection{Quantum Kernel SVM}
\begin{figure}[h]
	\centering
	\includegraphics
	[scale=0.7]{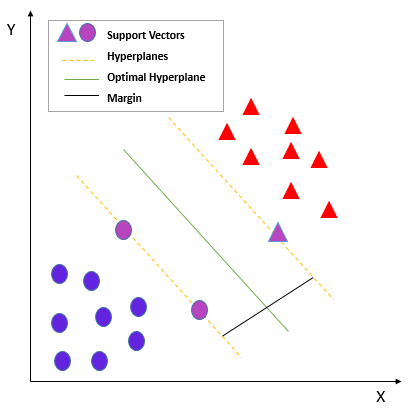} 
	\caption{linear SVM components.}
	\label{Fig:F2}
\end{figure}

The SVM is a widespread classifier model broadly utilized in medical applications. The binary SVM works to discover the best hyperplane, which classifies the features into two categories {+1, −1}. The SVM model works to great the margin between the best hyperplane and the nearest data points, called the support vectors (as shown in Figure \ref{Fig:F2}).
In 2019, two quantum algorithms in supervised ML are proposed \cite{havlivcek2019supervised}: variational quantum classifier (VQC) and quantum kernel estimator (QKE). In VQC and QKE, the quantum state space is used as the feature space. The VQC is based on a quantum variation circuit described in \cite{mitarai2018quantum,farhi2018classification} for data classification in a way like classical SVMs. The VQC consists of two stages (training and classification stage). In the training stage, the authors used four steps to compute the hyperplane between training data. The classification stage is used to classify new data with the correct label. The second technique, the main idea of QKE is to estimate a quantum kernel on quantum device and using this kernel with classical SVMs. The key idea of QKSVM is to apply a quantum kernel that would be hard to simulate with classical devices. Quantum kernels map data from classical $\vec{x}$ into quantum state $|\Phi(\vec{x})\rangle$.
\begin{equation}
K(\vec{x}, \vec{z})=|\langle\Phi(\vec{x}) \mid \Phi(\vec{z})\rangle|^{2}
\end{equation}
The steps of the hybrid QKSVM algorithms can be explained as follow
\begin{itemize}
\item The quantum computer is used to estimate the kernel ( applying the Equation \ref{Eq:quantumkernel}) from the training data of the all/informative genes of the Colon/Breast dataset.
\item The classical computer is used to obtain the support vectors based on the quantum kernel matrix.
\item The prediction stage is performed with the classical device according to the obtained support vectors based-quantum kernel from the training data.
\end{itemize}
The three quantum feature maps (kernels) are used with the proposed approach: ZFeaturemap, ZZFeatureMap, and PauliFeatureMap (a combination of $Z$ and $YY$ matrices). The quantum feature map function can be generated from the following equation:
\begin{equation}
\label{Eq:quantumkernel}
U_{\Phi(\vec{x})}=U_{\Phi(\vec{x})} H^{\otimes n} U_{\Phi(\vec{x})} H^{\otimes n}
\end{equation}
Where $\mu_{\Phi(\vec{x})}$ is the Pauli expansion unitary, $H$ is the Hadamard gate , and $n$ is the qubit number.
\begin{equation}
\mu_{\Phi(\vec{x})}=\exp \left(i \sum_{S \subseteq[n]} \phi_{S}(\vec{x}) \prod_{i \in S} P_{i}\right)
\end{equation}

\section{Results and Discussion}
\label{Sec:Results}
This section presents the used microarray colon and breast datasets and analyzes the experimental results acquired through the proposed approach. The results are implemented on Google Colab by using the Qiskit framework. Qiskit is a cross-platform Python library for quantum computation. The quantum feature map is executed on a quantum computer simulator with 20 qubits, called the QASM simulator. The results are presented in two parts for clarity.

\subsection{Microarray datasets}
\label{Subsec:data}
The proposed approach is evaluated using two microarray cancer datasets, namely Colon \cite{olshen2002deriving}, and Breast \cite{feltes2019cumida} datasets. The total number of genes in the Colon dataset is 2000.  There are 40 instances in the dataset that are assigned to the tumor and the other 22 instances are normal. Forty instances in the dataset are assigned to the tumor, and the other 22 instances are normal. The Breast dataset consists of 33,578 genes and 139 samples. Table 1 shows a description of the two used datasets and class distribution after using the SMOTE technique. Table \ref{Tbl:T1} shows a description of the two used datasets and class distribution after using SMOTE technique.
\begin{table}[h]
	\centering
	\caption{The description of the used datasets.}
	\label{Tbl:T1}%\resizebox{1\textwidth}{!}{%
		\begin{tabular}{ccccc} 			
		\hline
			\textbf{dataset}	& \textbf{Genes}  &\textbf{Samples} &\textbf{	Class Distribution}	& \textbf{Class Distribution after SMOTE}\\  \hline
			Colon&  2000& 62& 40 tumor 22 normal & 49 tumor  31 normal \\
			Breast& 33578 & 139 & 129 primary   10 normal & 129 primary  64 normal\\
			\hline
	\end{tabular}
	%}
\end{table}
\subsection{Evaluation Methods}
\label{Subsec:perf-measures}
The proposed approach performance is evaluated with five measures: accuracy, recall, precision, f1-measure, and receiver operating characteristic–area under curve (ROC–AUC) curve. The five measures can be calculated by a confusion matrix, which contains four elements (i.e., TP, TN, FP, and FN) defined as follows:
\begin{itemize}
	\item TP: true positive; indicates the proposed model correctly predicts cancer cases and is classified as cancer cases
	\item TN: true negative; means the proposed model correctly predicts normal and is classified as normal 
	\item FP: false positive; means the proposed model incorrectly predicts normal and is classified as normal
	\item FN: false negative; indicates the proposed approach incorrectly predicts cancer cases and is classified as a cancer case
\end{itemize}
\begin{equation}
\label{Eq:acc}
Accuracy (Acc.)= TP+TN / (TP+TN+FP+FN)
\end{equation}
\begin{equation}
\label{Eq:rec}
True-Positive Rate / Recall (Re.)= TN/TN+FP
\end{equation}
\begin{equation}
\label{Eq:Pre}
Precision (Pre.)= TP/TP+FP
\end{equation}
\begin{equation}
\label{Eq:F1}
F1-score = 2*(Pre. * Re.) / (Pre.+Re.)
\end{equation}
\begin{equation}
\label{Eq:FPR}
False-Positive Rate = FP/FP+TN 
\end{equation}
%%%%%%%%%%%%%%%%%%%%%%%%%%%%%%%%%%%%%%%%%%
\subsection{Results Analysis}
\label{Subsec:discussion}
\begin{figure}
	\centering
	\begin{subfigure}{.5\textwidth}
		\includegraphics[width=1\linewidth]{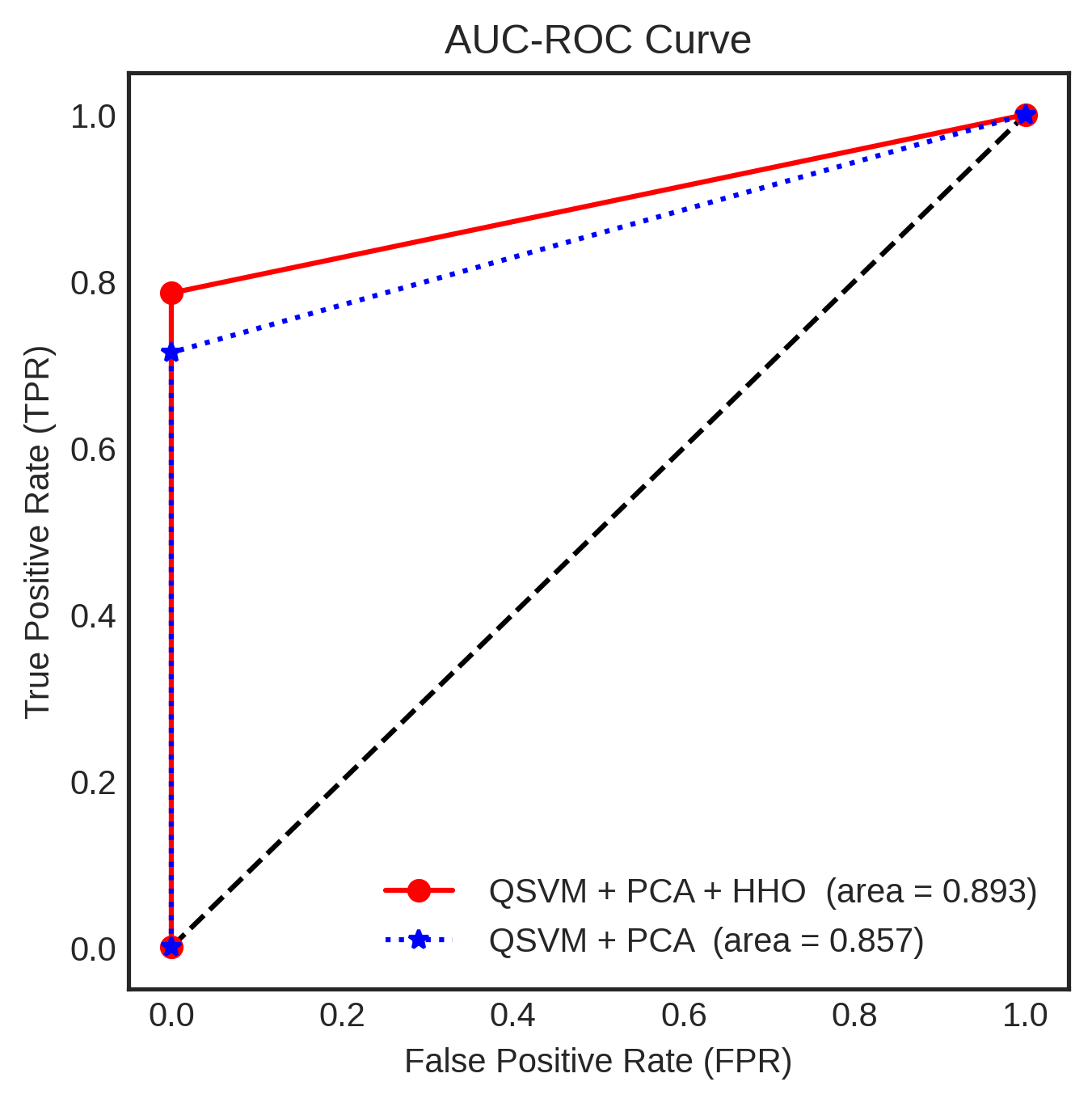}
		\caption{}
		\label{fig:F2-1}
	\end{subfigure}
	\begin{subfigure}{.49\textwidth}
		\centering
		\includegraphics[width=1\linewidth]{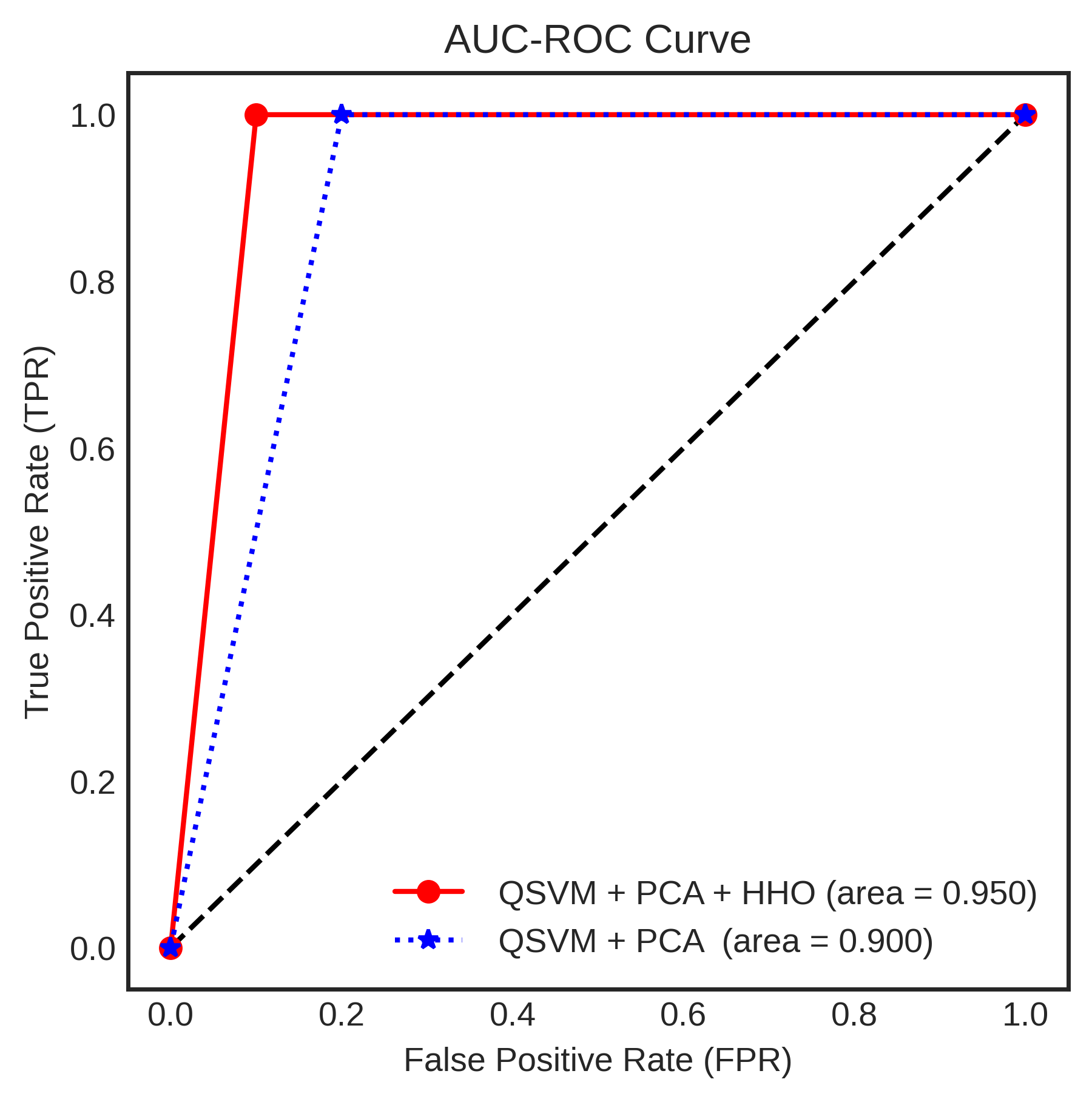}
		\caption{}
		\label{fig:F2-2}
	\end{subfigure}
	\caption{The AUC-ROC curve of the proposed approach and QKSVM+PCA for (a) Colon and (b) Breast datasets.}
	\label{fig:F2}
\end{figure}

\begin{figure}[]
	\centering
	\begin{subfigure}{.49\textwidth}
		\includegraphics[width=1\linewidth]{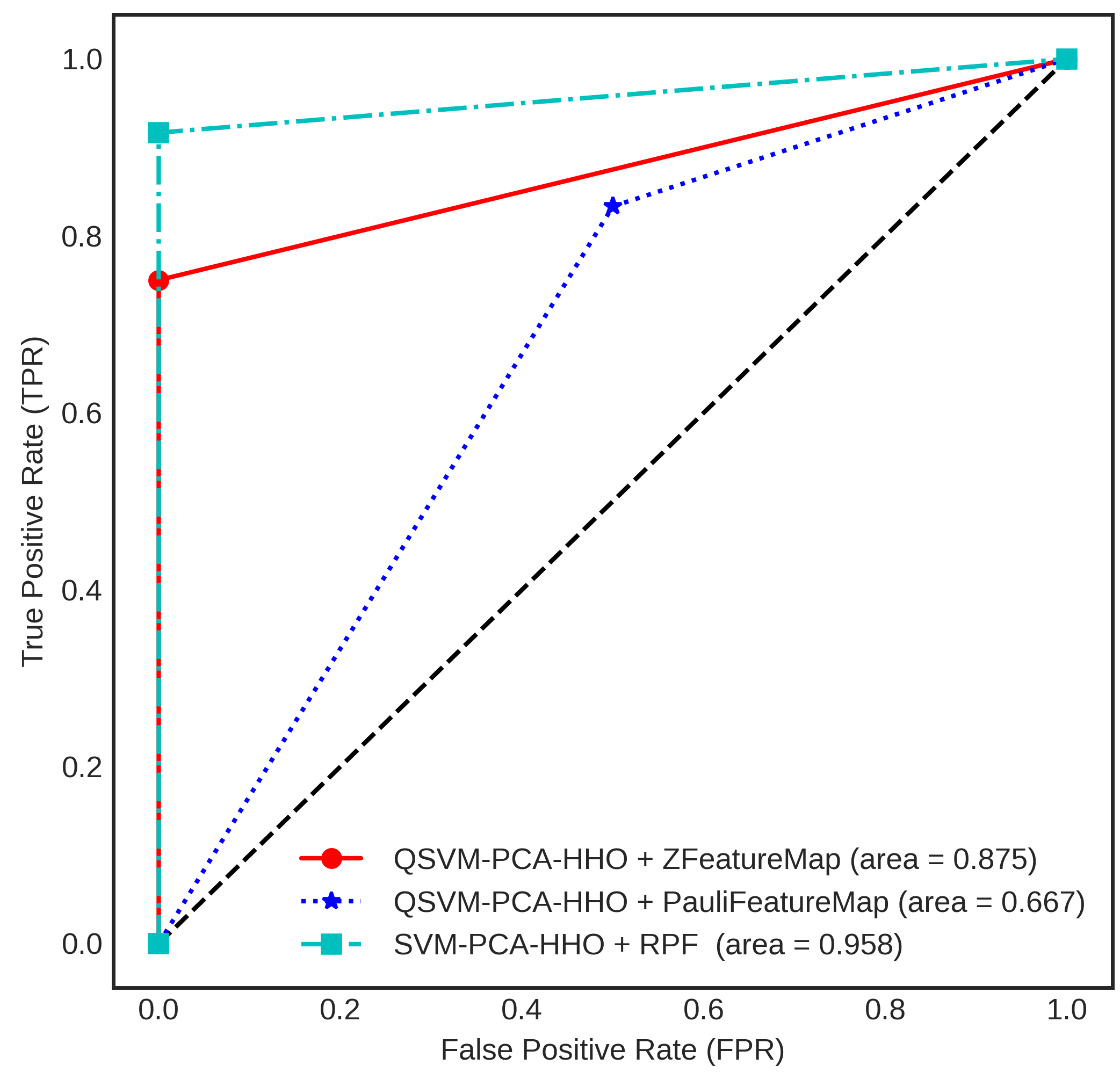}
		\caption{}
		\label{fig:F3-1}
	\end{subfigure}
	\begin{subfigure}{.49\textwidth}
		\centering
		\includegraphics[width=1\linewidth]{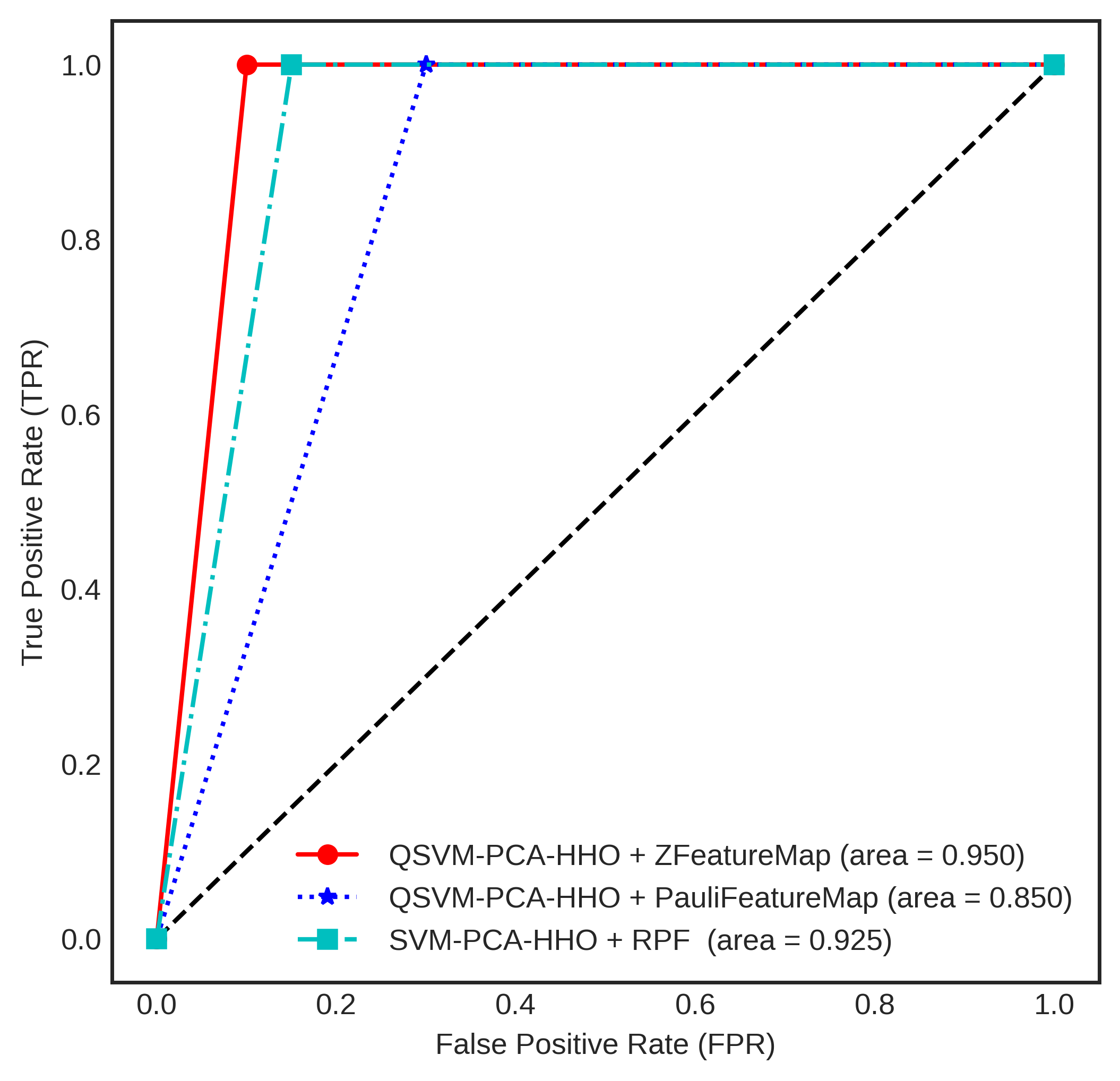}
		\caption{}
		\label{fig:F3-2}
	\end{subfigure}
	\caption{The AUC-ROC curve of the proposed approach with  different quantum kernel and classical kernel (RBF) (a) Colon and (b) Breast datasets.}
	\label{fig:F3}
\end{figure}

\begin{figure}
	\centering
	\includegraphics[scale=0.9]{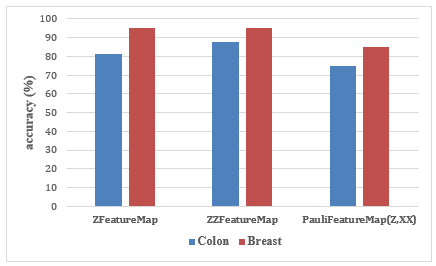} 
	\caption{Summarizes the accuracy measure of the two datasets with different quantum kernels.}
	\label{Fig:F4}
\end{figure}

Before the result analysis, Table \ref{Tbl:T2} shows the parameters setting of the quantum kernels used in this work. The parameters of QKSVM are feature map, the number of shots is 100,  the number of seed-simulator and seed-transpiler is 10,598. The first part of the results is applied with all genes and the selected genes by BHHO. The QKSVM algorithm is used with the ZZFeatureMap quantum kernel. First, the QKSVM algorithm with PCA is applied to two datasets with all genes. Finally, the selected genes by BHHO are used with the QKSVM-PCA. Table \ref{Tbl:T3} presents the classification performance on the test data set with all genes and the selected genes by BHHO. The proposed approach based on selected genes by BHHO enhances the overall performance of the colon and Breast datasets. In two data sets, the proposed model achieved better performance in terms of accuracy, recall, f1-score compared with that using all genes. The proposed model obtains the highest accuracies of 87.5\% and 95\% for the colon and breast datasets, respectively. Moreover, the proposed approach achieves the highest AUC-ROC score in the two datasets as shown in Figure \ref{fig:F2}.
\begin{table}[!htb]
	\centering
	\caption{parameter setting of quantum kernels}
	\label{Tbl:T2}%\resizebox{0.5\textwidth}{!}{%
		\begin{tabular}{cc}
			\hline
			\textbf{Feature Map}	& \textbf{Parameters}\\ \hline
		     ZZFeatureMap & \multirow{2}{*}{Dimensions-number=20, Depth=3, and Entanglement= linear}\\
		     PauliFeatureMap(Z,YY) \\
		    ZFeatureMap & Dimensions-number=20 , Depth=3\\
             
             \hline
	\end{tabular}
	%}
\end{table}
\begin{table}[!htb]
	\centering
	\caption{Classification performance based on all genes and the selected genes by BHHO}
	\label{Tbl:T3}%\resizebox{0.5\textwidth}{!}{%
		\begin{tabular}{cccccc}
			\hline
			\textbf{Dataset} &\textbf{Approach}& \textbf{ACC.(\%)}	& \textbf{Pre.(\%)}& \textbf{Re.(\%)}&  \textbf{F1-score(\%)}\\  \hline
		    \multirow{2}{*}{Colon}& QKSVM+PCA&  83.3 &	100	&71.4 &	83.3\\
		  & BHHO+PCA+QKSVM &	87.5 &	100 &	87.5 &	88 \\ \\
		     \multirow{2}{*}{Breast} &	QKSVM+PCA &	90 &	83.3 &	100	 &90\\
		    &BHHO+PCA+QKSVM &	95 &	90.9 &	100	 &95.2\\
		    \hline
	\end{tabular}
	%}
\end{table}
The second part offers the results of the proposed approach with another two quantum kernels (ZFeatureMap, and PauliFeatureMap) and classical kernel (RBF). Table \ref{Tbl:T4} presents the classification performance on the test data set with different functions. In the colon data, the classical SVM (RBF) outperformed the two kernel functions(ZFeatureMap and PauliFeatureMap). PauliFeatureMap is a combination of $Z$ and $YY$ Feature Maps. The SVM(RBF) obtains the highest accuracy of 93.7\%, whereas the QKSVM (PauliFeatureMap) obtains the lowest accuracy of 75\%. Also, the highest AUC is achieved by classical SVM as offered in Figure \ref{fig:F3}. In Breast data, the QKSVM (ZFeatureMap) achieves the highest accuracy with 95\%. By contrast, The QKSVM (PauliFeatureMap) obtains the lowest accuracy with 85\%. Besides, the highest accuracy is obtained by QKSVM (ZFeatureMap) as shown in Figure \ref{Fig:F4}. The quantum advantage is not noticed owing to the limited number of qubits and the current quantum hardware. The quantum advantage display in complex problems, in which classical computers solve Rebentrost et al.proposed a quantum version of the SVM for big data classification. Currently, this version does not run on quantum devices.

\begin{table}[!htb]
	\centering
	\caption{Classification performance based on different quantum kernels and classical RBF kernel.}
	\label{Tbl:T4}%\resizebox{0.5\textwidth}{!}{%
		\begin{tabular}{ccccccc}
			\hline
			\textbf{Dataset} &\textbf{Algorithm}&  \textbf{Kernel}& \textbf{ACC.(\%)}	& \textbf{Pre.(\%)}& \textbf{Re.(\%)}&  \textbf{F1-score(\%)}\\  \hline
		    \multirow{3}{*}{Colon} & \multirow{2}{*}{QKSVM}& ZFeatureMap	 &81.2 &	100 &	75	&85.7\\
		    &&PauliFeatureMap& 75 &	83.3 &	83.3 &	83.3\\
           &SVM	  &  RPF & 	93.7 &	100	 &91.6 &95.6 \\ \hline
           \multirow{3}{*}{Breast} & \multirow{2}{*}{QKSVM}& ZFeatureMap	 &95&	90 &	100	 &95.2\\
		    &&PauliFeatureMap& 85&	76.9&	100&86.9\\
           &SVM	  &  RPF & 92.5 &	86.9&	100 &	93\\
		    \hline
	\end{tabular}
\end{table}
\section{Conclusion and Future Work}
\label{Sec:Con}
Binary Harris hawks optimization is integrated with quantum kernel SVM for selecting the genes and classification of microarray datasets. The proposed approach aims to improve the microarray cancer classification performance based on the quantum kernel estimation with the informative genes. The two microarray datasets are applied for evaluating the proposed approach with all genes and the selected genes. The principal component analysis is applied to reduce the informative genes to match the qubit numbers of the quantum simulator. Various quantum feature maps are used with the QKSVM to convert data into quantum feature space. The proposed approach archives the higher accuracy with the selected genes 87.5\% and 95\% for colon and breast cancer, respectively. Also, the performance proposed approach is compared with different quantum kernels and classical kernel RBF. Due to the limited quantum hardware resources, the advantage of quantum SVM is not clear enough in real applications \cite{phillipson2020three,houssein2021hybrid}. In future work, more quantum kernels can be applied with real quantum devices. The variational quantum classifier can also be used with different datasets.

%\section*{Declaration of competing interest}
%The authors declare that there is no conflict of interest. The authors declare that they have no known competing financial interests or personal relationships that could have appeared to influence the work reported in this paper.

%\section*{Conflict of interest}
%The authors declare that there is no conflict of interest. %Non-financial competing interests. 

%\bibliographystyle{IEEEtran}
\footnotesize{\bibliography{References}}

\end{document}